\documentclass[10pt,a4paper,conference]{IEEEtran}

\usepackage{geometry}
\usepackage{subfigure}

\usepackage{amsmath,graphicx}
\usepackage{cite}  
\usepackage{url,subfigure}  
\usepackage{algorithm}
\usepackage{algorithmicx}
\usepackage{float}
\usepackage{bm}
\usepackage{multirow}
\usepackage{color}
\usepackage{xcolor}
\usepackage[bookmarks=true, colorlinks=true,hyperindex,breaklinks]{hyperref}

\newcommand{\PointDDD}{\mathbf{P}}

\newcommand{\IndexSet}{\mathcal{I}}
\newcommand{\List}{\mathcal{T}}
\newcommand{\RegionSet}{\mathcal{R}}
\newcommand{\Normal}{\mathbf{n}}
\newcommand{\Scale}{\mathbf{s}}
\newcommand{\PlaneDDD}{\mathbf{\Pi}}
\newcommand{\LineDDD}{\mathbf{L}}

\begin{document}
%
\title{Fast 3D Line Segment Detection From Unorganized Point Cloud}

\author{\IEEEauthorblockN{{Xiaohu~Lu$^1$, Yahui~Liu$^{2,3}$, Kai~Li$^4$}}
\IEEEauthorblockA{
$^1$Department of Civil, Environmental and Geodetic Engineering, Ohio State University, Columbus, USA \\
$^2$Department of Information Engineering and Computer Science, University of Trento, Trento, Italy \\
$^3$Fondazione Bruno Kessler, Trento, Italy \\
$^4$Department of Electrical and Computer Engineering, Northeastern University, Boston, USA \\
\text{\{lu.2037@osu.edu, yahui.liu@unitn.it, kaili@ece.neu.edu\}}} }


%


\maketitle

\begin{abstract}
This paper presents a very simple but efficient algorithm for 3D line segment detection from large scale unorganized point cloud. Unlike traditional methods which usually extract 3D edge points first and then link them to fit for 3D line segments, we propose a very simple 3D line segment detection algorithm based on point cloud segmentation and 2D line detection. Given the input unorganized point cloud, three steps are performed to detect 3D line segments.
Firstly, the point cloud is segmented into 3D planes via region growing and region merging. Secondly, for each 3D plane, all the points belonging to it are projected onto the plane itself to form a 2D image, which is followed by 2D contour extraction and Least Square Fitting to get the 2D line segments. Those 2D line segments are then re-projected onto the 3D plane to get the corresponding 3D line segments. Finally, a post-processing procedure is proposed to eliminate outliers and merge adjacent 3D line segments. Experiments on several public datasets demonstrate the efficiency and robustness of our method. More results and the C++ source code of the proposed algorithm are publicly available at \url{https://github.com/xiaohulugo/3DLineDetection}.
\end{abstract}

%
\IEEEpeerreviewmaketitle

\section{Introduction}
\label{Sec:Introduction}

Line segment is a good representation of the scene, especially in the urban where exits lots of man-made structures with regular shape. In the passing decade, line segment detection from 2D image has been well studied and several efficient algorithms have been proposed~\cite{Von2010LSD,Almazan2017MCMLSD,Lu2015Cannylines} including the widely used LSD~\cite{Von2010LSD}. However, there are still not enough works addressing 3D line segment detection, which can generally be classified into three categories: point based, plane based and image based. 

The point based methods usually detect the boundary points first and then fit for the 3D line segment via Least Square Fitting method. Given the input point cloud, we can get the boundary points via convex hull method~\cite{Lin2017Facet,Awrangjeb2016Using} or classify the whole point cloud into boundary/non-boundary categories via designed features. The original convex hull algorithm is efficient to find the boundary points of a convex but not suitable for extracting concave outlines. \cite{Jarvis1977Computing} improved the convex hull, which allowed some concavities in the extracted shape. However, the convex hull based method is still ``not found to be particularly useful due to an uneven point distribution in the point cloud"~\cite{Awrangjeb2016Using}. On the other hand, there are various kinds of features for boundary/non-boundary point cloud classification, including the curvature or surface variation~\cite{Bazazian2015Fast}, the Gauss Map~\cite{Weber2010Sharp}, the multi-scale normal difference~\cite{Ioannou2012Difference}, and the features based on eigenvalues and eigenvector~\cite{Hackel2016Contour}. In the work of~\cite{Ioannou2012Difference}, the normal of each point is calculated in different scales, then the ones with higher normal difference are classified as boundary points. In the recent work~\cite{Hackel2016Contour}, a contour score for each individual point is predicted with a binary classifier using a set of features extracted from the point's neighbourhood. Given the extracted boundary point, we can build an octree and then fit for 3D line segments via region growing and Least Square Fitting. The biggest problem of these point based methods lies on the feature point itself, which are not robust to noise. 

The plane based methods are also reasonable choices because a 3D line segment can also be considered as the intersection of two 3D planes. However, the plane based methods are more often used in building modelling from airborne LiDAR~\cite{Sampath2010Segmentation}, where the points are usually not very dense. For example, in the work of~\cite{Sampath2010Segmentation}, planar points are firstly distinguished from the non-planar ones, and then clustered into individual roof segments based on their normals, finally the adjacent segments are detected and intersected to from the vertexes and boundary line segments. In another work~\cite{Moghadam2013Line}, 3D line segments extracted from point cloud via plane intersection are combined with their 2D corresponding line segments to calibrate the extrinsic parameters of range and image sensors. One common drawback of the plane based methods is that the terminals of the intersection line are generally difficult to be determined, besides, this kind of methods are not suitable for small planes.

The image based methods convert the input point cloud into image(s) first, and then apply line segment detector to extract 2D line segments on the image(s), finally these 2D line segments are re-project to the point cloud to get the final 3D line segments. Converting the point cloud into image is a strategy often used in point cloud processing, for example, in the recent work~\cite{Ge2017Automatic}, the point cloud is segmented into several cross-sections, then the point cloud in each cross-section is projected onto a specific plane and then extracted feature points from different 2D images. In another recent work~\cite{Lin2015Line}, the input point cloud is converted into a collection of projected images with different viewpoints which are evenly placed on a sphere surrounding the point cloud. Then the LSD algorithm~\cite{Von2010LSD} is applied to extract 2D line segments on each of these projected images, which are then back-projected into the original point cloud to get the initial 3D line segments. The key problem of these image based methods lies on the image generating strategy, the number, the resolution and the view points of these projected images are hard to be determined.

The proposed method belongs to the image based category. To overcome the shortages of traditional image based methods, we proposed an simple but efficient 3D line segment detection method which contains the following three steps:
\begin{itemize}
	\item \emph{Point Cloud Segmentation}: the input point cloud is segmented into 3D planes via region growing and region merging.
	\item \emph{Plane Based 3D Line Detection}: for each 3D plane, all the points belonging to it are projected onto the plane itself to form a 2D image, which is followed by 2D contours extraction and Least Square Fitting to get the 2D line segments for each plane. Those 2D line segments are then re-projected onto the 3D plane to get the 3D line segments.
	\item \emph{Post-processing}: outliers of the 3D planes and 3D line segments are removed by exploring the structural information of the scene, after that a line merging procedure is performed to merge adjacent 3D line segments.	
\end{itemize}

\section{Algorithm}
\label{Sec:Algorithm}

\subsection{Point Cloud Segmentation}
\label{Sec:Point Cloud Segmentation}

Segmentation is one of the most important pre-processing step for automatic processing of point clouds, which has been well studied in the passing decades. Generally, the proposed point cloud segmentation methods can be classified into three main categories: edge/border based, region growing based and hybrid (refer to~\cite{Grilli2017Review} for a review). To get planar segmentations from the point cloud, traditional methods like~\cite{Rabbani2006Segmentation} which perform region growing simply on point-level are too sensitive to parameters. A much robuster way is extracting small compact regions first and then merge those regions to get the final result. Based on this observation, our method takes three steps to segment the input point cloud into planes: normal calculation, region growing and region merging.

\textbf{\textit{Normal Calculation}}: Given a set of unorganized point clouds, we employ the K Nearest Neighbour (KNN) method to find the neighbours of each data point and estimate the normal of the neighbouring surface via the Principal Component Analysis (PCA)~\cite{Lu2016Pairwise}. Firstly, we build a k-d tree by applying the nanoflann library~\cite{Blanco2014Nanoflann}. Then, for each point, we can calculate its covariance matrix via the following equation:
	\begin{equation}\label{Eq:CovarianceMatrix}
		\bm{\Sigma} = \frac{1}{k} \sum\nolimits_{i=1}^{k} ( \PointDDD_{i} - \overline{\PointDDD} ) ( \PointDDD_{i} - \overline{\PointDDD} )^T, 
	\end{equation} 
where $\bm{\Sigma}$ denotes the $3\times3$ covariance matrix, $\PointDDD_{i}$ is a neighbouring data point, and $\overline{\PointDDD}$ represents the mean vector of the KNNs. Solving the following standard eigenvalue equation via Singular Value Decomposition (SVD), we can finally get the normal of the point, which is the third eigenvector in the matrix of eigenvectors $\mathbf{V}$.
	\begin{equation}\label{Eq:EigenvalueMatrix}
		\bm{\lambda} \mathbf{V} = \bm{\Sigma} \mathbf{V} 
	\end{equation}
where $\mathbf{V}$ is the matrix of eigenvectors and $\bm{\lambda}$ is the matrix of eigenvalues. What should be noticed is that the third value $\lambda_3$ of $\bm{\lambda}$ can represent the curvature of the point, the smaller the value of $\lambda_3$ is, the smaller the curvature of the point will be. Besides, since the KNNs of $\PointDDD_{i}$ are obtained, we can briefly estimate the scale of the point distribution in the neighbourhood. Simply, we define the scale of $\PointDDD_{i}$ as the distance between $\PointDDD_{i}$ and its third closest neighbouring point.
We denote the normal, the curvature, the scale and the set of KNNs of point $\PointDDD_{i}$ as $\Normal_{\PointDDD_{i}}$, $\lambda_{\PointDDD_{i}}$, $\Scale_{\PointDDD_{i}}$ and $\IndexSet_{\PointDDD_{i}}$, respectively.

\begin{figure}
   	\centering
	\begin{tabular}{c}
		\includegraphics[width=1\linewidth]{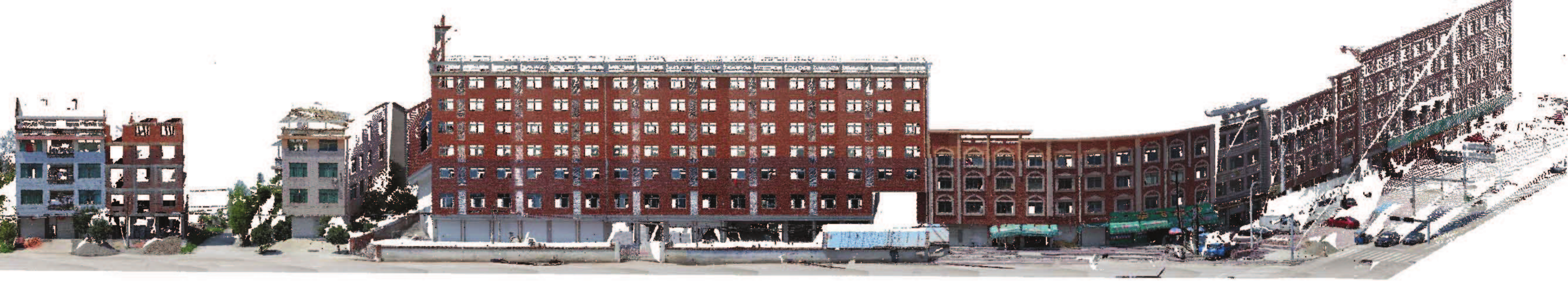} \\
		(a) input point cloud (with RGB for better view) \\
		\includegraphics[width=1\linewidth]{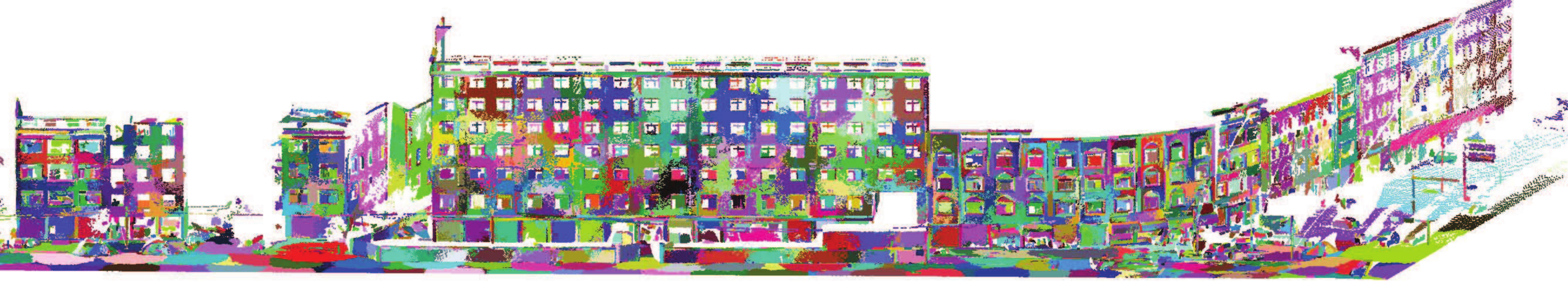} \\
		(b) segmentation result before merging \\
		\includegraphics[width=1\linewidth]{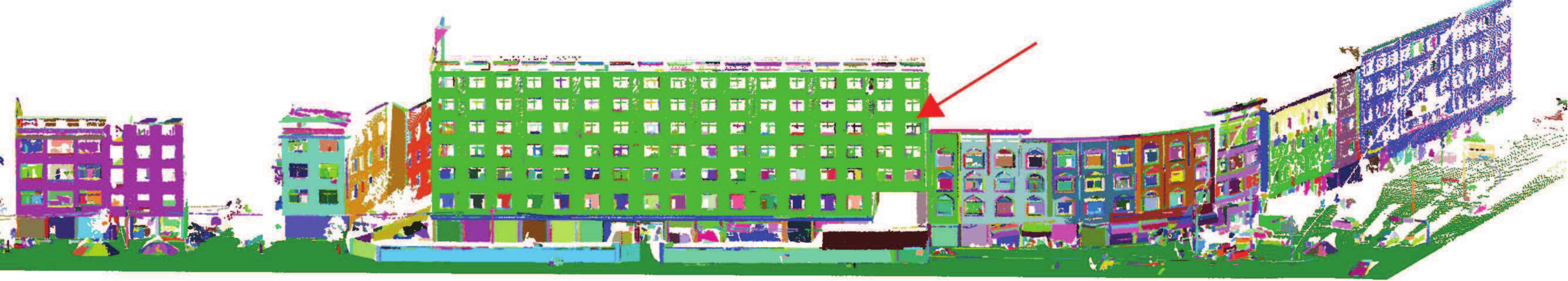} \\			
		(c) segmentation result after merging		
	\end{tabular}   	
   	\caption{A demonstration of the proposed point cloud segmentation algorithm on a vehicle-mounted point cloud.}
   	\label{Fig:demo_segmentation}
\end{figure} 

\textbf{\textit{Region Growing}}: Given $\{ \Normal_{\PointDDD_{i}}, \lambda_{\PointDDD_{i}}, \Scale_{\PointDDD_{i}}, \IndexSet_{\PointDDD_{i}} \}$ of each point $\PointDDD_{i}$, the region growing procedure aims to extract all the planes in the input point cloud, which is performed as follows: 
(1) All the points are sorted ascendantly according to their curvature. 
(2) Starting from the foremost unprocessed point $\PointDDD_{s}$ with the smallest curvature, we create a list $\List$ to store all the coplanar points with $\PointDDD_{s}$ being the first one. For each unprocessed point $\PointDDD_{i}$ in $\List$, we traverse each of the unprocessed point $\PointDDD_{j}$ in its neighbourhood $\IndexSet_{\PointDDD_{i}}$ and add $\PointDDD_{j}$ to list $\List$ if the following conditions are satisfied:
	\begin{equation}\label{Eq:GrowConditions}
	\begin{aligned}
	  & \arccos( \left| \Normal_{\PointDDD_{s}} ^ \top \cdot \Normal_{\PointDDD_{j}} \right| ) < \theta, \\
	  & \left| \Normal_{\PointDDD_{s}} ^ \top \cdot \overrightarrow{\PointDDD_{s}\PointDDD_{j}} \right| < th_{o},\\
	  & || \overrightarrow{\PointDDD_{s}\PointDDD_{j}} || < th_{p},	  
	\end{aligned}
	\end{equation}
where $\theta = 15^{\circ}$ is a constant angle threshold, $th_{o} = \Scale_{\PointDDD_{s}}$ is the orthogonal distance threshold and $th_{p} = 50 \Scale_{\PointDDD_{s}}$ is the parallel distance threshold. Both $th_{o}$ and $th_{p}$ are self-adapted by the scale of $\PointDDD_{s}$. The first and the second condition make sure that $\PointDDD_{j}$ is coplanar with $\PointDDD_{s}$, while the third condition ensures the region is compactly around $\PointDDD_{s}$. Once all the points in $\IndexSet_{\PointDDD_{i}}$ are traversed, we mark $\PointDDD_{i}$ as processed. The region is growing point by point until all the points in list $\List$ are processed, the result of which is a planar region with all its points stored in list $\List$. 
(3) The second step is iteratively performed until all the points are processed. The set of all the planar regions obtained is denoted as $\RegionSet$.

\begin{figure*}
   	\centering
   	\renewcommand{\tabcolsep}{1pt}	
	\begin{tabular}{ccc}
		\includegraphics[width=0.32\linewidth]{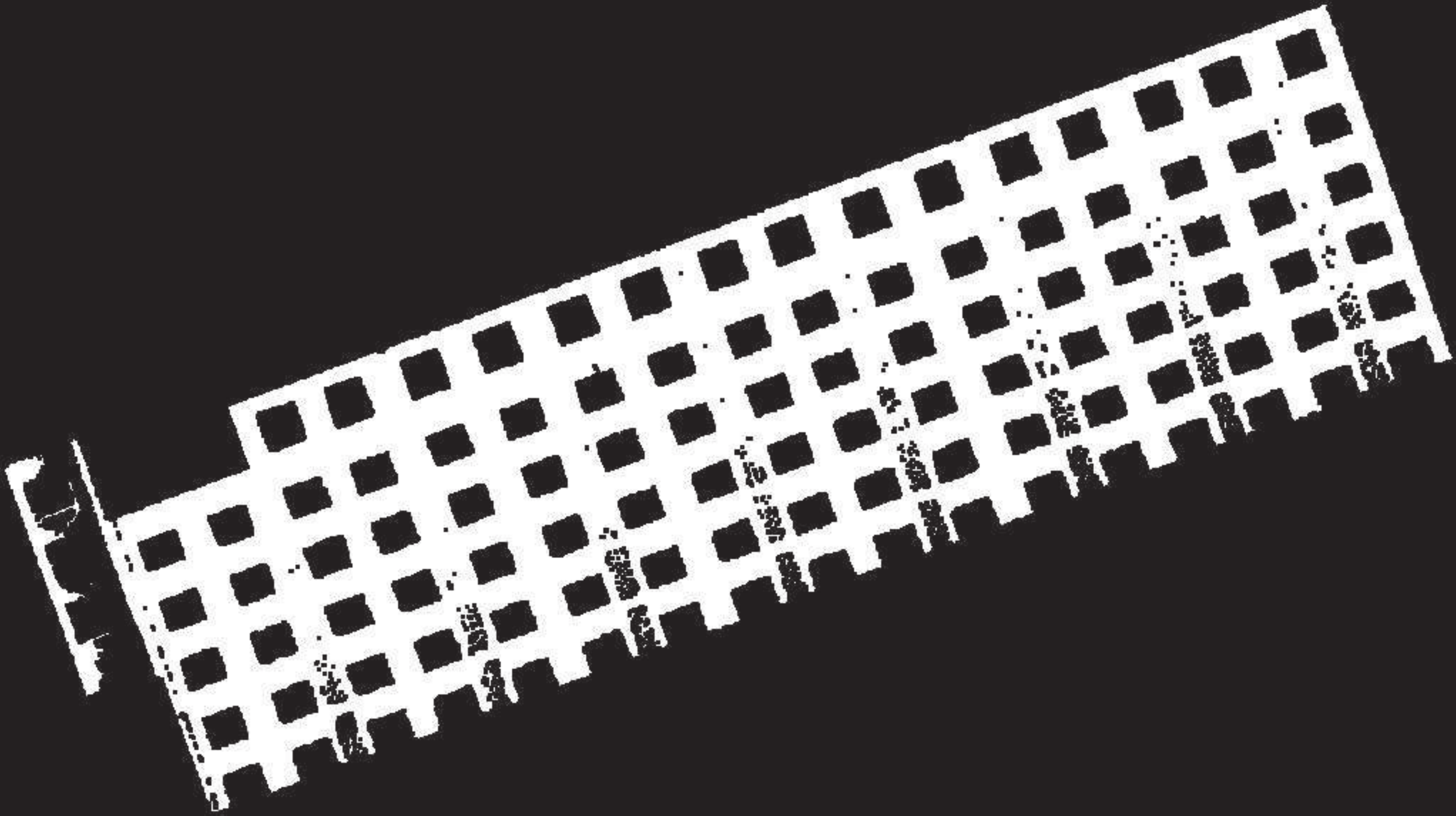} & 
		\includegraphics[width=0.32\linewidth]{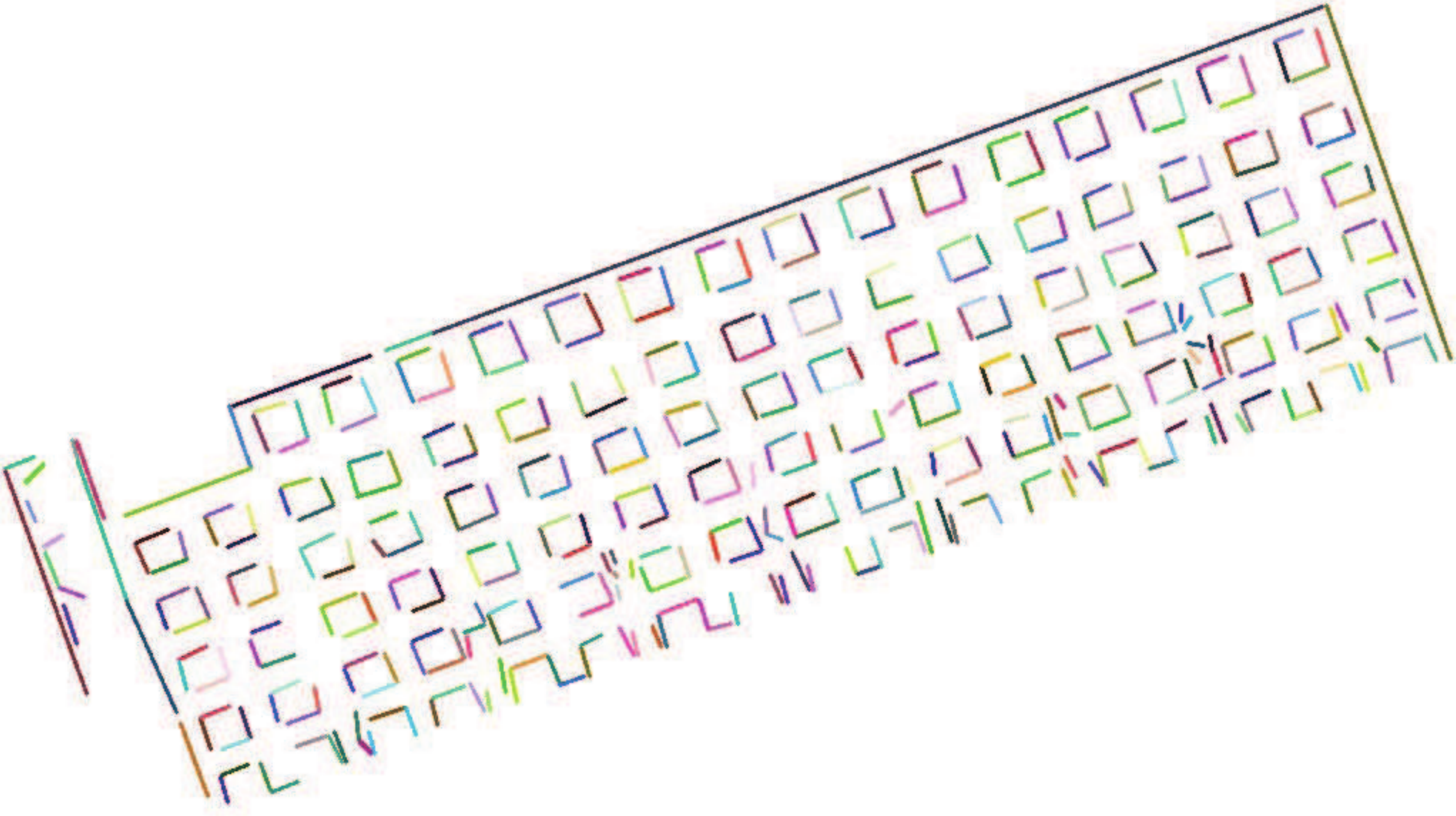} & 
		\includegraphics[width=0.32\linewidth]{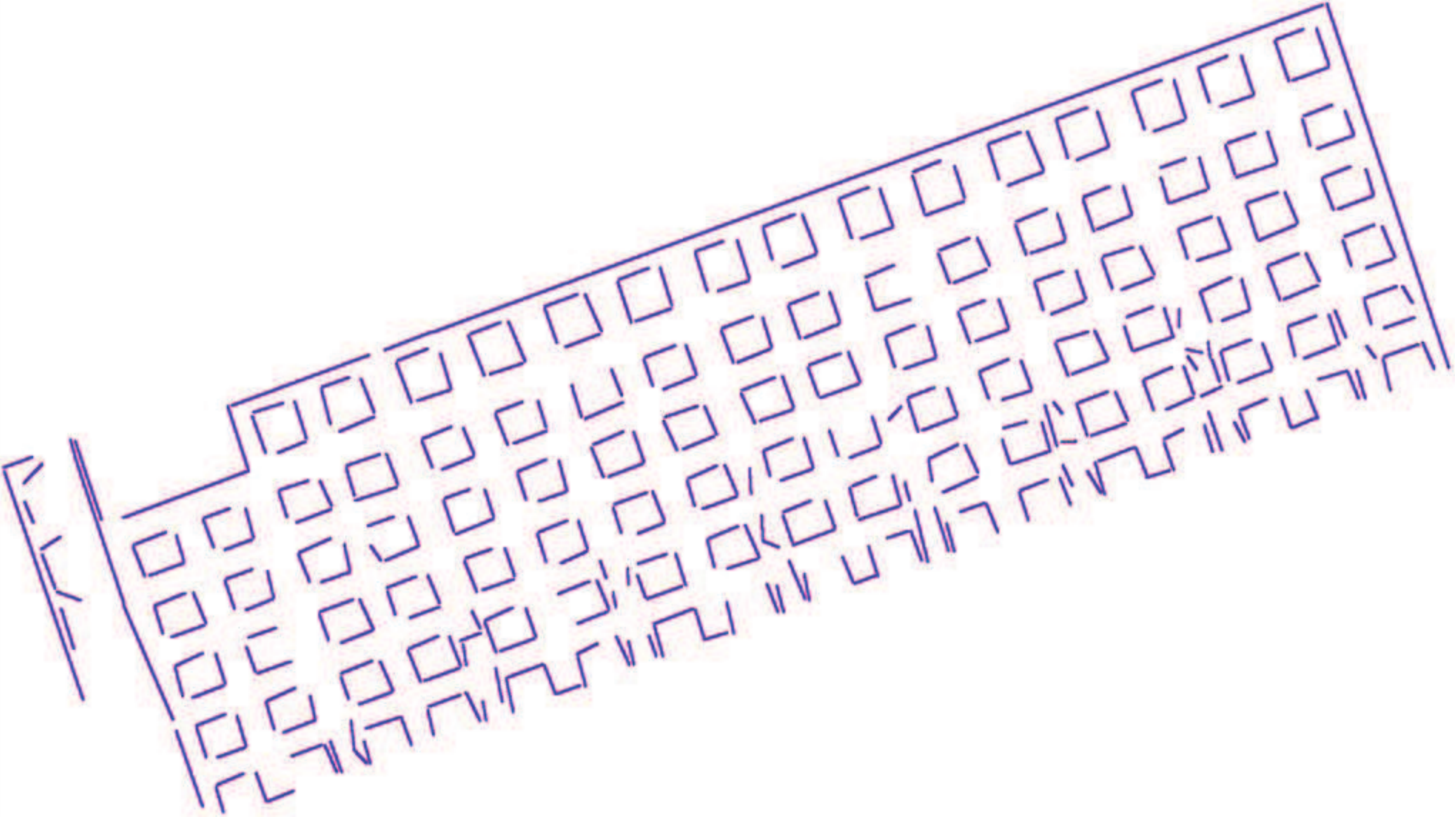}\\
		(a) 3D-2D projection  &(b) 2D line detection   & (c) 2D-3D re-projection\\	
	\end{tabular}   	
   	\caption{A demonstration of the proposed image based 3D line detection algorithm on an exemplary 3D plane (marked with red arrow in Figure~\ref{Fig:demo_segmentation} (c)).}
   	\label{Fig:demo_line_detection}
\end{figure*} 

\textbf{\textit{Region Merging}}: The proposed region merging strategy contains the following three steps.
Firstly, for each region $\RegionSet_{i}$, we fit a plane to it using all its points in the same way as introduced in normal calculation, the normal, curvature and scale of $\RegionSet_{i}$ are denoted as $\Normal_{\RegionSet_{i}}$, $\lambda_{\RegionSet_{i}}$ and $\Scale_{\RegionSet_{i}}$, respectively. Also, for each point in the point cloud, we assign a label to it, which is the index of the region it belongs to.
After that, we traverse each region to obtain its neighbouring regions. For each point $\PointDDD_{m}$ in a region $\RegionSet_{i}$, we traverse all the points in its neighbourhood $\IndexSet_{\PointDDD_{m}}$. If there is a point $\PointDDD_{n}$ in $\IndexSet_{\PointDDD_{m}}$ whose label is different to that of $\PointDDD_{m}$, we consider that $\PointDDD_{n}$ is a boundary point and take the plane that $\PointDDD_{n}$ belongs to as an adjacent region of $\RegionSet_{i}$. In this way, for each region in $\RegionSet$, we can get its adjacent ones.
Finally, similar to the region growing procedure introduced above, we try the same way to aggregate the coplanar regions, which differs from the previous region growing procedure in the following two aspects: (1) the adjacent regions are used instead of the neighbourhood of each point; (2) the third condition is no longer used. After that, we fit a plane for each merged region in the same way as that introduced in the first step.

The result of the point cloud segmentation procedure is a set of planes, in which the normal $\Normal_{\PlaneDDD}$ and the belonging points $\List_{\PlaneDDD}$ of each plane are all obtained.

\subsection{Plane Based 3D Line Detection}
\label{Sec:Plane Based 3D Line Detection}

Given the extracted 3D planes, we perform three steps ( as Figure~\ref{Fig:demo_line_detection} shows) to extract the 3D line segments. (1) \emph{3D-2D Projection}: for each plane, its belonging points are orthographically projected onto the plane itself, which is then converted into a binary image via a gridding procedure with auto-adaptive parameters determined by the plane itself. (2) \emph{2D Line Detection}: contours are extracted from the binary image first and then line segments are fitted via Least Square Fitting. (3) \emph{2D-3D Re-projection}: the detected 2D line segments are re-projected onto the 3D plane to get the corresponding 3D line segments.

\subsubsection{3D-2D Projection}
\label{Sec:3D-2D Projection}

\begin{figure}
   	\centering
	\begin{tabular}{c}
		\includegraphics[width=0.95\linewidth]{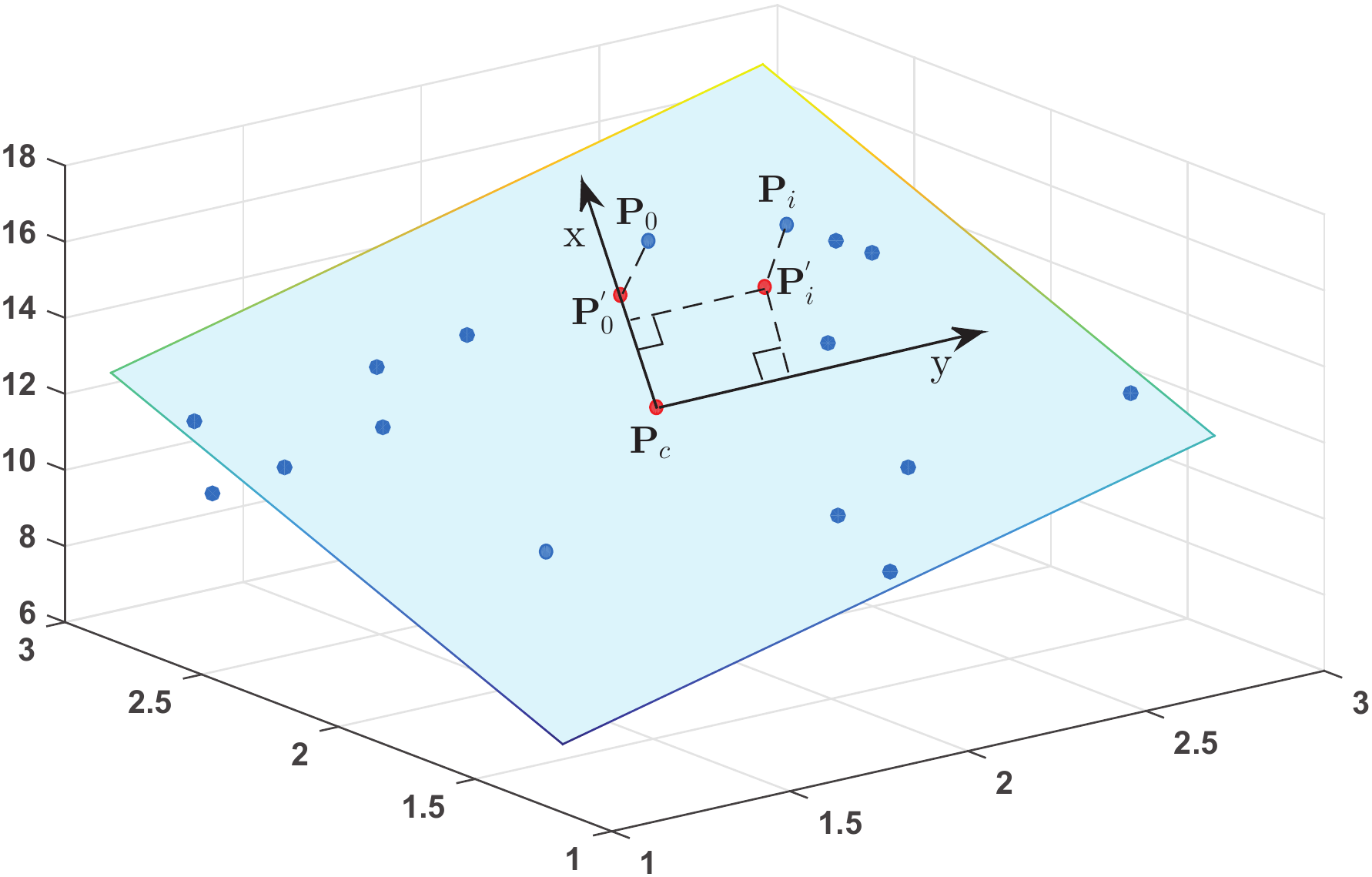}
	\end{tabular}   	
   	\caption{Demonstration of the procedure which converts a 3D point to 2D.}
   	\label{Fig:3Dto2D}
\end{figure}

For a 3D plane $\PlaneDDD$ obtained in the above procedure, we aim to convert it from 3D to 2D in order to form a binary image, from which 2D line segments can be efficiently detected. The 3D-2D projection is conducted as follows, which is also demonstrated in Figure~\ref{Fig:3Dto2D}.

Firstly, the central point $\PointDDD_{c}$ of the plane $\PlaneDDD$ is calculated as the average of all its belonging points. Then we project the first point (denoted as $\PointDDD_{0}$) in $\List_{\PlaneDDD}$ (the belonging points of plane $\PlaneDDD$) onto the plane with the corresponding projected point denoted as $\PointDDD^{'}_{0}$. After that we set the $\overrightarrow{\PointDDD_{c}\PointDDD^{'}_{0}}$ as the x-axis, with the positive direction from $\PointDDD_{c}$ to $\PointDDD^{'}_{0}$. The positive direction of x-axis is denoted as $v_{x}$ and that of the y-axis can be calculate as $v_{y} = v_{x} \times \Normal_{\PlaneDDD}$. With the central point $\PointDDD_{c}$, the positive directions of both x-axis and y-axis, the 2D plane coordinate $(x_i,y_i)$ of a point $\PointDDD_{i}$ can be easily calculate as follows:

	\begin{equation}\label{Eq:3Dto2D}
	\begin{aligned}
	  & \overrightarrow{\PointDDD_{c}\PointDDD^{'}_{i}} = \overrightarrow{\PointDDD_{c}\PointDDD_{i}} - (\overrightarrow{\PointDDD_{c}\PointDDD_{i}} \ \ \cdot \ \ \Normal_{\PlaneDDD})  \Normal_{\PlaneDDD}, \\
	  & x_i = \overrightarrow{\PointDDD_{c}\PointDDD^{'}_{i}} \ \ \cdot \ \ v_{x}, \\
	  & y_i = \overrightarrow{\PointDDD_{c}\PointDDD^{'}_{i}} \ \ \cdot \ \ v_{y}.
	\end{aligned}
	\end{equation}

For each 3D point in $\List_{\PlaneDDD}$, we can calculate its 2D plane coordinates via Equation~\ref{Eq:3Dto2D}. We can also get the scale $\Scale_{\PlaneDDD}$ of the plane, which is empirically defined as $0.75 \{ \Scale_{\PointDDD} \}_{0.9}$, where $\{ \Scale_{\PointDDD} \}$ is the set of ascendantly sorted
scales of points in $\List_{\PlaneDDD}$ and $\{ \Scale_{\PointDDD} \}_{0.9}$ denotes the $90\%_{th}$ value in the set. Given the set of 2D plane points $\{ (x_i,y_i) \}$ and the scale of the plane, we can then convert these points into a binary image via gridding, which is performed as follows.

For a plane $\PlaneDDD$, firstly, the minimal and maximal values of the x and y coordinates of all the 2D plane points are obtained, which are denoted as $x_{min}, x_{max}, y_{min}$ and $y_{max}$, respectively. Then a binary image is created with its width equals $[(x_{max} - x_{min})/\Scale_{\PlaneDDD}]+1$, its height equals $[(y_{max} - y_{min})/\Scale_{\PlaneDDD}]+1$. And the pixel coordinate $(u_i,v_i)$ of each 2D plane point $(x_i,y_i)$ can be obtained as follows:
	\begin{equation}\label{Eq:xy2uv}
	\begin{aligned}
	  & u_i = [(x_i - x_{min}) / \Scale_{\PlaneDDD}], \\
	  & v_i = [(y_i - y_{min}) / \Scale_{\PlaneDDD}].	  
	\end{aligned}
	\end{equation}
We set the grey value as 255 for the pixels converted from the 2D plane points, and those of the rest pixels as 0. In this way, we achieve the goal of converting a 3D plane into a 2D binary image. In real application, there may exit some small holes on the original binary image, which is due to the fact that the distances between some points may be larger than $\Scale_{\PlaneDDD}$. Increasing the value of $\Scale_{\PlaneDDD}$ can largely eliminate those holes, however, it will reduce the resolution of the binary image. After numerous experiments, we find that a more reasonable way is applying a dilation and then a erosion with a default $3 \times 3$ kernel on the original binary image. Figure~\ref{Fig:demo_line_detection} (a) shows the result after a dilation and a erosion, which not only keeps the resolution of the original image and also makes up most holes.

\subsubsection{2D Line Detection}
\label{Sec:2D Line Detection}
Given the binary image like Figure~\ref{Fig:demo_line_detection} (a), we can then easily extract 2D line segments from it using line segment detectors like LSD~\cite{Von2010LSD} and CannyLines~\cite{Lu2015Cannylines}. Both of them can lead to good line segment detection results, however, in practice, we found that contour detection followed by a least square line fitting procedure is the most flexible way. First, the tolerance for line fitting can be adjusted. Besides, the topological information between line segments on a plane can be kept, which is very useful for the following post-processing to eliminate outliers of line segments. Thus, we detect 2D line segments via the following steps:

Firstly, contours are found via calling the function $\textit{findContours}$ in OpenCV. Then those contours whose sizes are less than 40 pixels are removed for the reason that they are not likely to contain useful line segments. After that, each useful contour is divided into straight segments in the same way as introduced in \cite{Lu2015Cannylines}, which are then fitted via Least Square Fitting. Figure~\ref{Fig:demo_line_detection} (b) shows the line segment detection result of Figure~\ref{Fig:demo_line_detection} (a), in which different line segments are marked in different colours. 

\subsubsection{2D-3D Re-projection}
\label{Sec:2D-3D Projection}
Given the 2D line segments extraction from the binary image, we can then get the corresponding 3D line segments via 2D-3D re-projection, which is the inverse process of the 3D-2D projection and contains the following two steps: (1) From image pixel to 2D plane point: for each line segment whose terminal coordinates are in pixel $(u_i,v_i)$, we convert them into 2D plane coordinates $(x_i,y_i)$ via the inverse process of Equation~\ref{Eq:xy2uv}. (2) From 2D plane point to 3D point: for each 2D plane point $(x_i,y_i)$, we calculate its corresponding 3D point $\PointDDD_i$ via the inverse process of Equation~\ref{Eq:3Dto2D}. In this way, we can obtain all the 3D line segments on a 3D plane as Figure~\ref{Fig:3Dto2D} (c) shows. Performing this procedure for each 3D plane, finally, we can get all the 3D line segments in the input point cloud.

\subsection{Post-processing}
\label{Sec:Post-processing}
In the above procedures, to make sure that no useful scene information is missing, only the short contours are eliminated in Section~\ref{Sec:2D Line Detection}. As a result, the initial 3D line segments obtained contain many outliers. To remove those outliers and make the detection result completer and cleaner, an robust and efficient post-processing procedure is introduced in this section.

As we have introduced above, for 2D line detection, we extract contours first and then fit line segments from each contour instead of detecting line segments directly from the input image. This strategy keeps the topological informations of line segments, that's to say we can obtain a hierarchical relationship from plane $\rightarrow$ contours $\rightarrow$ lines. Based on this hierarchical relationship, we propose our post-processing strategy which is much robuster than those methods that process on individual line segment. Our post-processing contains two steps: outliers removal and line merging.

\textbf{\textit{Outliers Removal}}: Based on the observation that inlier line segments often show structural information like parallelism and orthogonality, we propose to utilize the local structural information to remove outliers, which is performed as follow.
(1) Firstly, we remove outlier 3D planes. For each 3D plane $\PlaneDDD$, all its belonging 3D line segments are sorted descendantly according to their length and the longest one is considered as the referring line of the first cluster. Then, starting from the foremost un-clustered line segment, each un-clustered 3D line segment (denoted as $\LineDDD_i$ ) is considered belonging to the cluster whose referring line has an orientation within $10^{\circ}$ deviation to that of $\LineDDD_i$. If the orientation deviations between $\LineDDD_i$ and any referring line are all larger than $30^{\circ}$, we create a new cluster and take $\LineDDD_i$ as its referring line. This procedure is iteratively conducted until all the 3D lines on $\PlaneDDD$ are processed. The sum of lines' lengths of each cluster is calculated and then utilized to descendantly sort these clusters. Generally, if the 3D plane is a regular one, like building facades and windows, there will be strong structural information like parallelism and orthogonality, while those outlier planes, like trees and vegetables, usually don't show such structural information. Thus, we propose a criteria to briefly judge if a 3D plane is outlier:
	\begin{equation}\label{Eq:outlier planes}
	\begin{aligned}
	l(c_1) + l(c_2) < 0.3*l(all)
	\end{aligned}
	\end{equation}
where $l(c_1)$ and $l(c_2)$ is the sum of line lengths of the foremost and the second cluster, respectively, and $l(all)$ is the length of all the 3D line segments in $\PlaneDDD$. What should be noticed is that the orientations of the foremost and the second cluster are defined as the structural orientations of $\PlaneDDD$.

(2) Then, we remove outliers of 3D line segments via the contour to which they belong. Our strategy is based on the observation that if the contour is structural then the length threshold $th_{l}$ to discard the outliers should be small, otherwise a large length threshold should be used. Thus, for a plane $\PlaneDDD$, firstly, we traverse all its belonging 3D line segments to find out the structural ones, which are defined as those whose orientations are within a tolerance ($10^{\circ}$ is used) to that of any one of the structural orientations of $\PlaneDDD$. Then the structural line ratio $t$ of the contour is defined as the length of the structural lines dividing the that of all the lines on the contour. Given the structural line ratio $t$ and the scale ($\Scale_{\PlaneDDD}$) of the plane, our strategy is as follows: 
\begin{itemize}
	\item if $t>75\%$, $th_{l} = 10 \Scale_{\PlaneDDD}$;
	\item if $75\% \geq t \geq 50\%$, $th_{l} = 20 \Scale_{\PlaneDDD}$;
	\item if $t < 50\%$, $th_{l} = 40 \Scale_{\PlaneDDD}$.
\end{itemize}

\begin{table*}[!htp]
	\caption{Computing results in eight public point clouds.}
	\vspace{-0.5em}
	\renewcommand{\tabcolsep}{3pt}
	\renewcommand\arraystretch{1.2}
	\centering
	\begin{tabular}{c|c|c|c|c|c|c|c|c}
		\hline
		\multicolumn{1}{c|}{Data set} & \multicolumn{1}{c|}{Point cloud} & \multicolumn{1}{c|}{Points number} & \multicolumn{1}{c|}{Lines number} 
		& \multicolumn{1}{c|}{Planes number} & \multicolumn{1}{c|}{Segmentation (s)} & \multicolumn{1}{c|}{Line detection (s)} & \multicolumn{1}{c|}{Post-processing (s)} & \multicolumn{1}{c}{Total time (s)}\\
		\hline
		\multirow{5}{*}{Semantic3D} 
		& Bildstein1 & 29,697,591  & 4856  & 2596  & 110.276  & 20.452  & 0.327  & 131.055 \\
		& Bildstein3 & 23,995,481  & 3543  & 2256  & 91.466   & 17.3    & 0.172  & 108.938 \\
		& Bildstein5 & 24,919,498  & 5001  & 2545  & 95.257   & 19.281  & 0.234  & 114.772 \\
		& Stgallen1  & 31,179,769  & 10268 & 5722  & 112.508  & 28.392  & 1.294  & 142.194 \\
		& Stgallen3  & 31,643,853  & 11052 & 6495  & 102.945  & 32.557  & 1.638  & 137.14 \\		
		\hline  
		\multirow{3}{*}{Shape Repertory} 
		& Castle & 32,754,893  & 12376 & 9752  & 94.365  & 51.277  & 5.382  & 151.024\\
		& Church & 7,632,351   & 2031  & 1176  & 23.057  & 10.483  & 0.078  & 33.618 \\
		& Facade & 29,465,648  & 12925 & 10362 & 91.385  & 37.44   & 6.177  & 135.002 \\	
		\hline 		                                             
	\end{tabular}\label{Tab:result_public_dataset}
\end{table*} 

\textbf{\textit{Line Merging}}: The line merging procedure is proposed to merge 3D line segments that are too close to each other, so the final result will be more cleaner. There line merging procedure is performed as follow. 
Firstly, the latitudes of each inlier of 3D line segment is calculate as $asin(z)$, where $z$ is the z-component of the line's normalized orientation. Then a histogram is built with a bin step as $6^{\circ}$ for the latitudes of all the inliers. After that, all the inliers are sorted descendantly according to their length. Starting from the foremost unprocessed 3D line segment $\LineDDD_{i}$, we traverse each 3D line segment $\LineDDD_{j}$ in the bin of $\LineDDD_{i}$ along with its left bin and right bin to find the line hypotheses which meet the following criteria:
	\begin{equation}\label{Eq:merging hyp}
	\begin{aligned}
	| d_{\LineDDD_{i}} - d_{\LineDDD_{j}} |/mag < 0.1,
	\end{aligned}
	\end{equation}
where $d_{\LineDDD_{i}}$ and $d_{\LineDDD_{j}}$ are the distances from the original point to $\LineDDD_{i}$ and $\LineDDD_{j}$, respectively, $mag$ is the magnitude of the coordinates of the input point cloud, namely $mag = ||\PointDDD_{0}||$, which is the distance between the first point of the input point cloud to the original point. By dividing $mag$, we can make sure that $| d_{\LineDDD_{i}} - d_{\LineDDD_{j}} |/mag$ will not increase proportionally with the magnitude of the coordinates. $0.1$ is an empirical value which is big enough for judging close line segments. 
Then, for each merging hypothesis, we calculate the perpendicular distances of its two terminals to $\LineDDD_{i}$ and discard it if any of these two distances is larger than $4 \Scale_{\PlaneDDD}$, where $\Scale_{\PlaneDDD}$ is the scale of the 3D plane $\LineDDD_{i}$ belongs to. For each 3D line segment $\LineDDD_{j}$ who is close enough to $\LineDDD_{i}$, if those two 3D line segments are overlapped or the gap between them is small enough (we set the gap threshold as $10 \Scale_{\PlaneDDD}$), we then update the terminals of $\LineDDD_{i}$ with the farthest two among the four terminals which are made up of the two terminals of $\LineDDD_{i}$ and the two by projecting $\LineDDD_{j}$'s terminals on $\LineDDD_{i}$ and mark $\LineDDD_{j}$ as merged. What should be noticed is that we choose to extend $\LineDDD_{i}$ instead of fitting a new 3D line segment for the reason that the fitted line may deviate to a location where exits no real 3D line segment in the scene.
Finally, all the unmerged 3D line segments are output as the final detection result.

\section{Experimental Results}
\label{Sec:Experimental Results}

\textbf{\textit{Parameters Setting}}: As we can see from above sections, there are generally two categories of parameters in our algorithm: angle-related and distance-related. The angle-related parameters, like $\theta$ in region growing and orientation deviation threshold in outliers removal, are suitable for most applications, which require no tuning. As to the distance-related parameters, like $th_{o}$ and $th_{p}$ in region growing and $th_{l}$ in line merging, there is also no need for them to be adjusted because the scales for the planes and lines are estimated beforehand. Thus, here we give a general setting of parameters: $\theta = 15^{\circ}$, $th_{o} = \Scale_{\PointDDD_{s}}$ and $th_{p} = 50 \Scale_{\PointDDD_{s}}$, where $\Scale_{\PointDDD_{s}}$ is the scale of 3D point $\PointDDD_{s}$ and $\Scale_{\PlaneDDD}$ is the scale of 3D plane $\PlaneDDD$. In all the following experiments, which are performed on a PC with Intel Core i5-3350 3.1GHz CPU and 16GB RAM, we applied this setting without adjustment.

\begin{figure}
   	\centering
	\begin{tabular}{c}
		\includegraphics[width=0.95\linewidth]{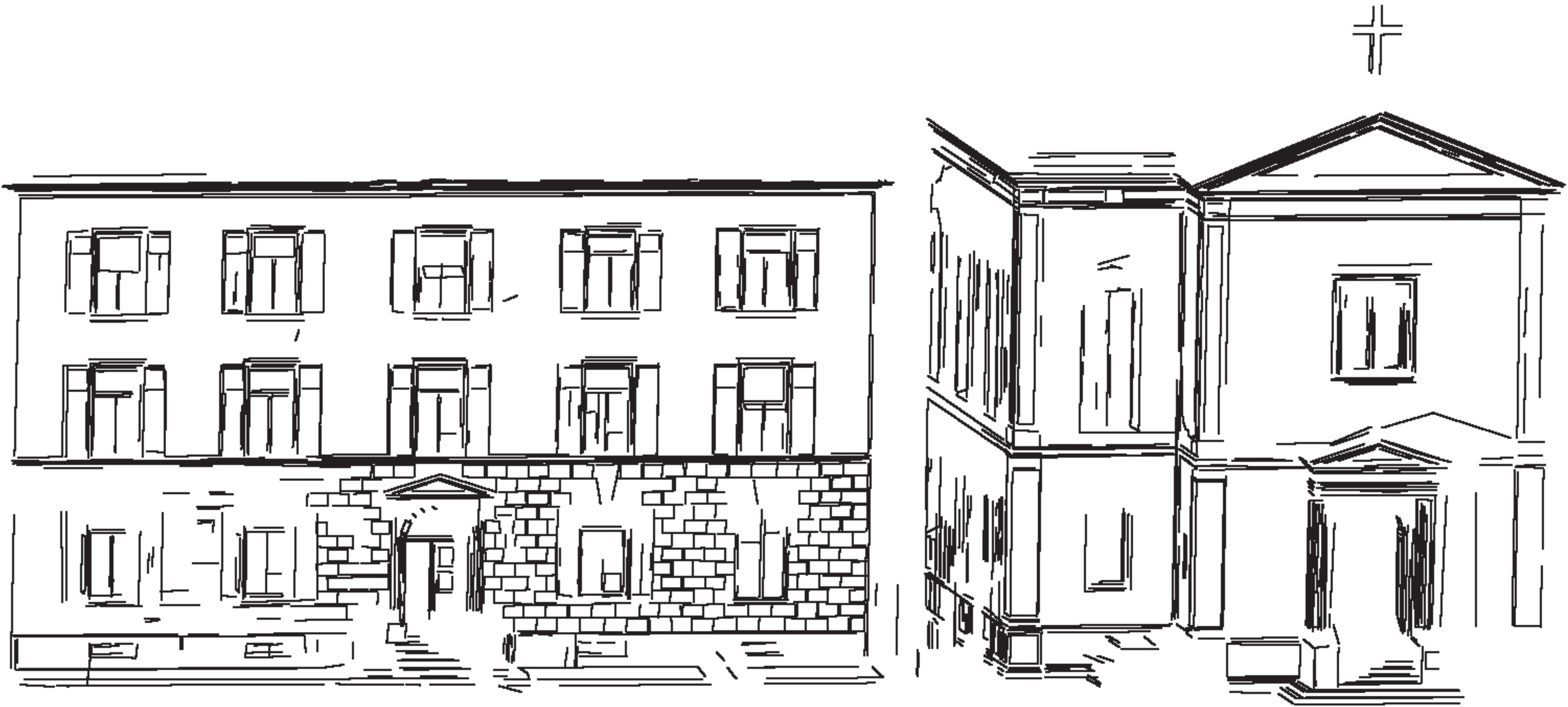}
	\end{tabular}   	
   	\caption{Details of the 3D line detection result of our method on Stgallen3.}
   	\label{Fig:result_demo_chruch}
\end{figure}

\begin{figure*}[!htp]
   	\centering
	\begin{tabular}{c}
		\includegraphics[width=0.9\linewidth]{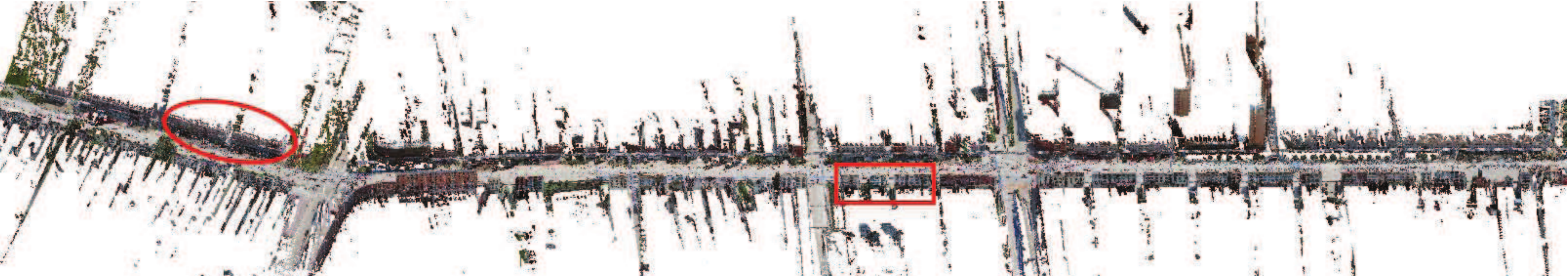} \\
		\includegraphics[width=0.9\linewidth]{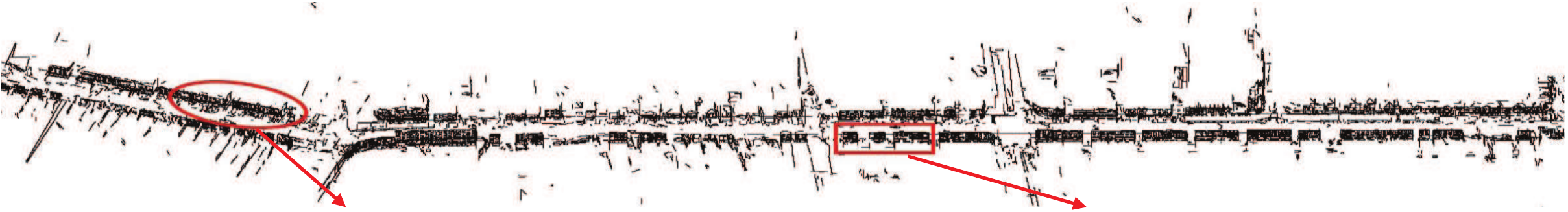} \\
	\end{tabular}
	\begin{tabular}{cc}
		\includegraphics[height=0.1\linewidth]{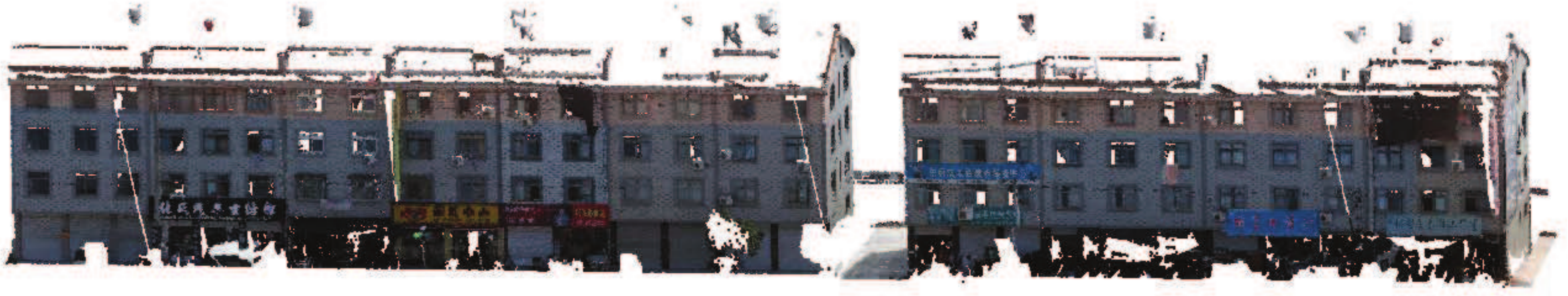}  & 
		\includegraphics[height=0.1\linewidth]{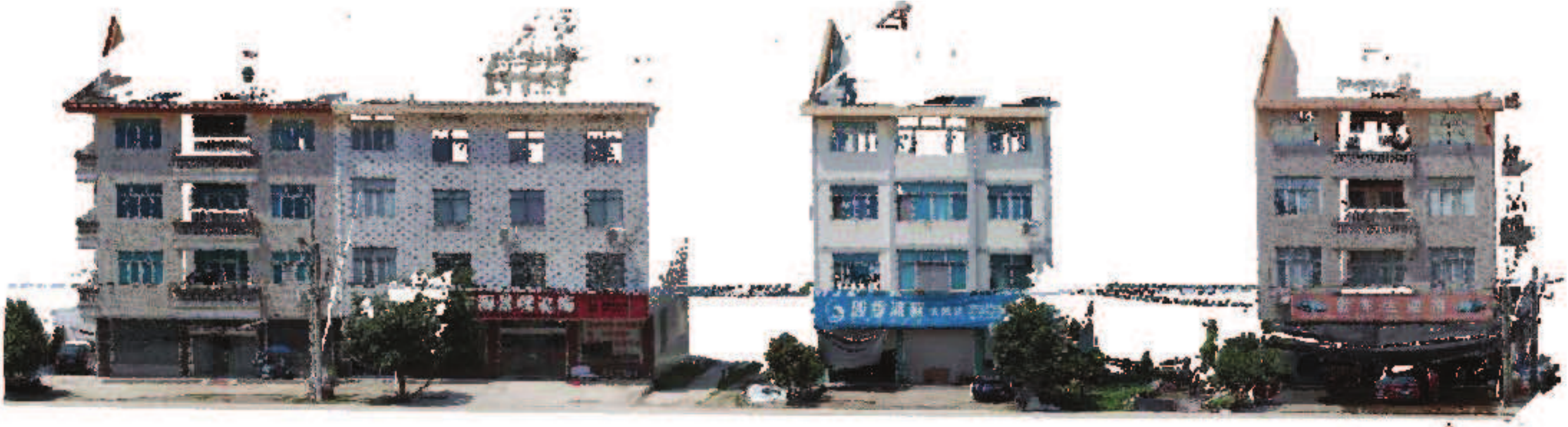}\\
        	        \includegraphics[height=0.1\linewidth]{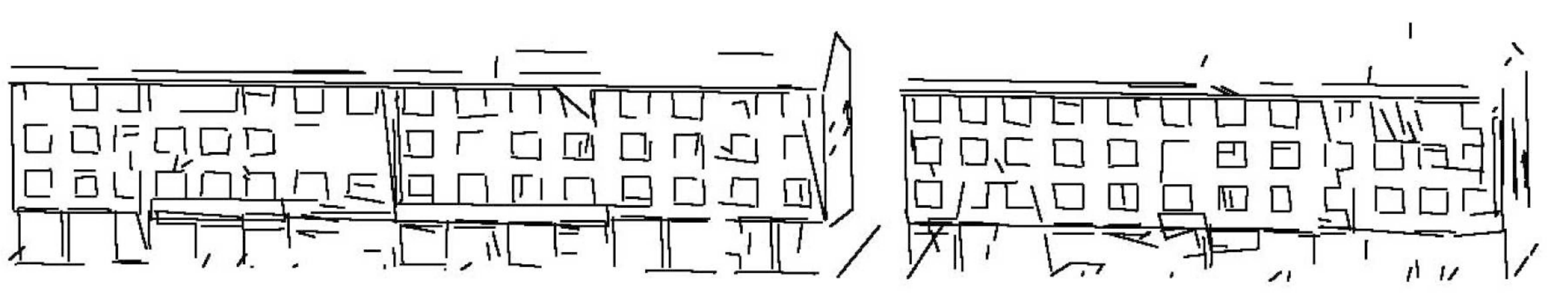}  & 
	        \includegraphics[height=0.1\linewidth]{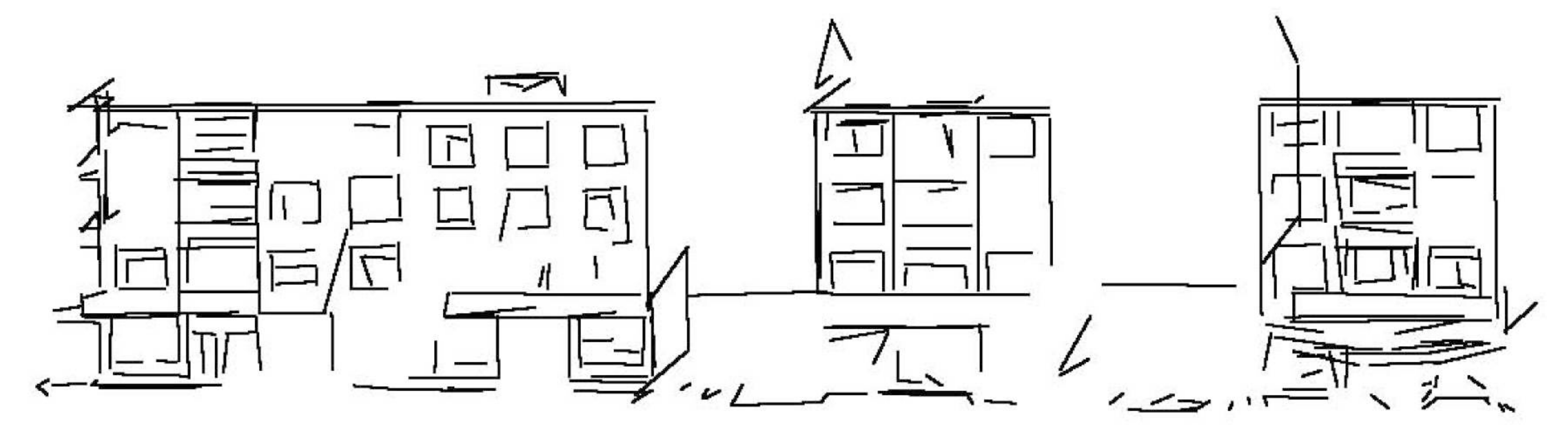}\\
	\end{tabular}	   	
   	\caption{3D line segment detection results of our method on a street-view point cloud with 22,003,946 points. Totally 13679 3D line segments and 5815 3D plans are detected by our method in 98.983s.}
   	\label{Fig:result_demo_street}
\end{figure*}

\textbf{\textit{Performance on Public Dataset}}: To evaluate the performance of the proposed 3D line segment detection method, we tested it on two public point cloud datasets: Semantic3D~\footnote{Available at \url{http://www.semantic3d.net/}} and Shape Repertory~\footnote{Available at \url{http://visionair.ge.imati.cnr.it/ontologies/shapes/}}. Eight point clouds were utilized, among which 5 were from Semantic3D and 3 were from the Shape Repertory. Table~\ref{Tab:result_public_dataset} demonstrates the statistic results of our method on those point clouds, from which we can get the following observations: most of the computing time is costed by segmentation of point cloud; the 3D line detection procedure is quite efficient (counts only 1/6 of total computing time); the computing time of post-processing is usually negligible. Figure~\ref{Fig:result_demo_chruch} shows the local details of the 3D line detection result on Stgallen3, we can see that not only the large structures, like the frames of the facades, but also the small details, like the windows and blocks on the left build, are well recovered. What should be noticed is that the short 3D line segments of the blocks are usually removed as outliers by traditional methods. However, in our post-processing procedure, they are kept because they are structural lines of the blocks, which demonstrates the effectiveness of our post-processing strategy. Besides these 8 point clouds, we have also tested our method on plentiful datasets. Figure~\ref{Fig:result_demo_street} shows the detection results of our method on a street-view point cloud with 22003946 points, 13679 lines segments and 5815 plans are detected by our method in only 98.983s. We can see from Figure~\ref{Fig:result_demo_street} that most of the details of the buildings along the street have been recovered very well, which demonstrate the potential of our method on vectorization of street view point cloud (more results are available at our github repository~\footnote{Available at \url{https://github.com/xiaohulugo/3DLineDetection}}). Based on the results on plentiful datasets, we found that the computing time of our method for point clouds with 30M, 20M, 10M, 5M, 2M and 1M points approximates to 130s, 80s, 40s, 20s, 8s and 4s, respectively.

\begin{table}[t]
	\caption{Comparison between our method and Lin et al.'s method~\cite{Lin2017Facet} in three public point clouds. $N_{l}$, $t_{s}$, $t_{l}$ and $t_{all}$ denote the number of 3D line segments, the time for segmentation, the time for 3D line detection and the total time, respectively.}
	\vspace{-0.5em}
	\renewcommand\arraystretch{1.2}
	\renewcommand{\tabcolsep}{2pt}
	\centering
	\begin{tabular}{c|cccc|cccc}
		\hline
		\multirow{2}{*}{\shortstack{Point cloud (size)}} & \multicolumn{4}{c|}{Lin et al.~\cite{Lin2017Facet}}  & \multicolumn{4}{c}{Ours} \\
		& $N_{l}$  & $t_{s}$ & $t_{l}$ & $t_{all}$  & $N_{l}$  & $t_{s}$ & $t_{l}$ & $t_{all}$ \\
		\hline
		Bildstein1(2M)   & 1104  & 19.5  & 17.6  & \textcolor{red}{37.1}    & 993    & 4.9   & 3.8   & \textcolor{red}{8.7} \\					
		\hline		
		Castle(32M)      & 21412 & 296.8 & 246.2 & \textcolor{red}{543.0}   & 12376  & 95.4  & 54.7  & \textcolor{red}{150.1}\\
		\hline		
		StSulpice(6M)    & 8816  & 80.9  & 37.7  & \textcolor{red}{118.6}   & 5450   & 22.1  & 8.0   & \textcolor{red}{30.1} \\				
		\hline	   			                                             
	\end{tabular}\label{Tab:comprison}
\end{table} 

\begin{figure*}[!htp]
   	\centering
	\begin{tabular}{ccc}
		\includegraphics[height=0.3\linewidth]{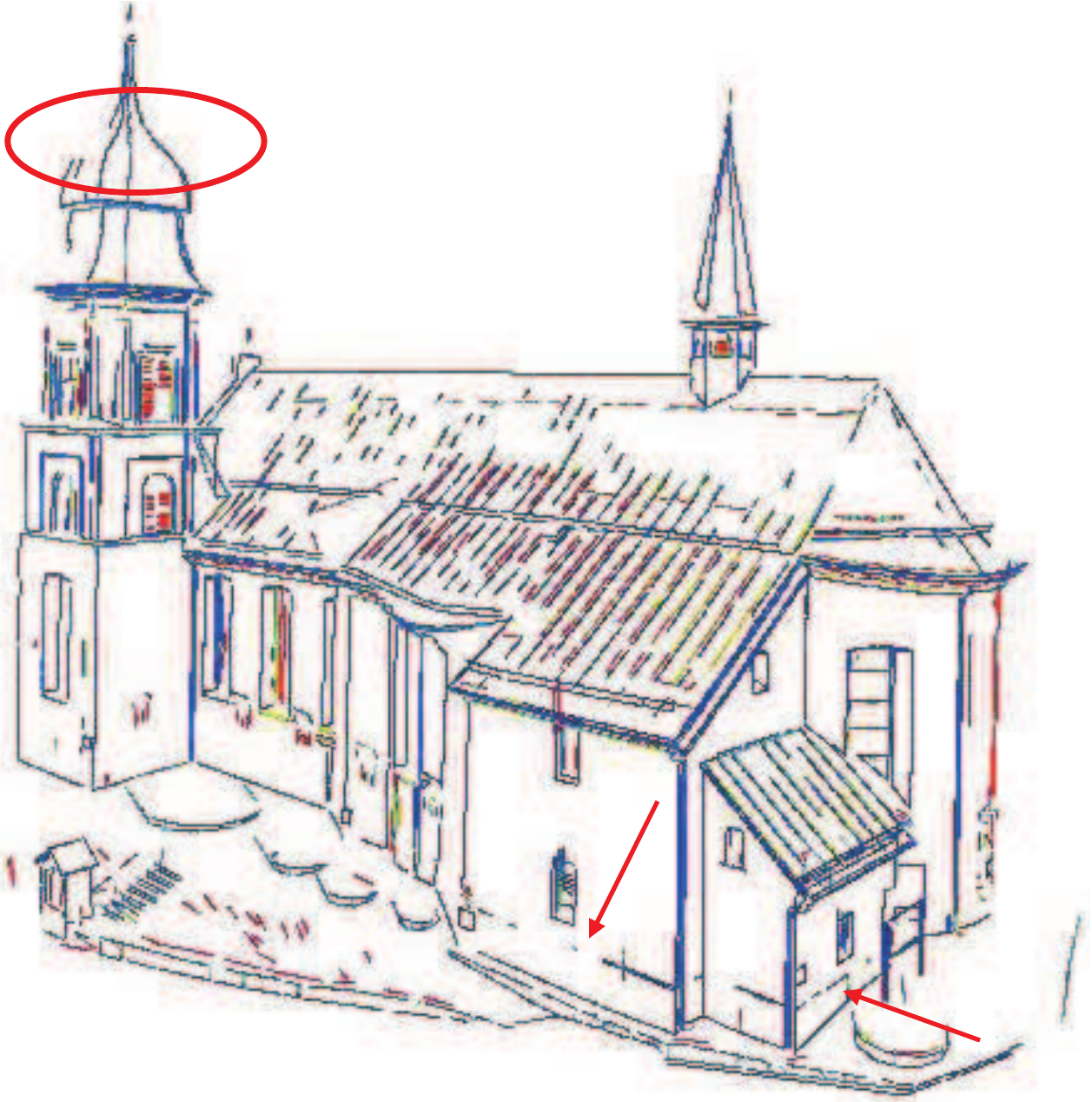} & 
		\includegraphics[height=0.3\linewidth]{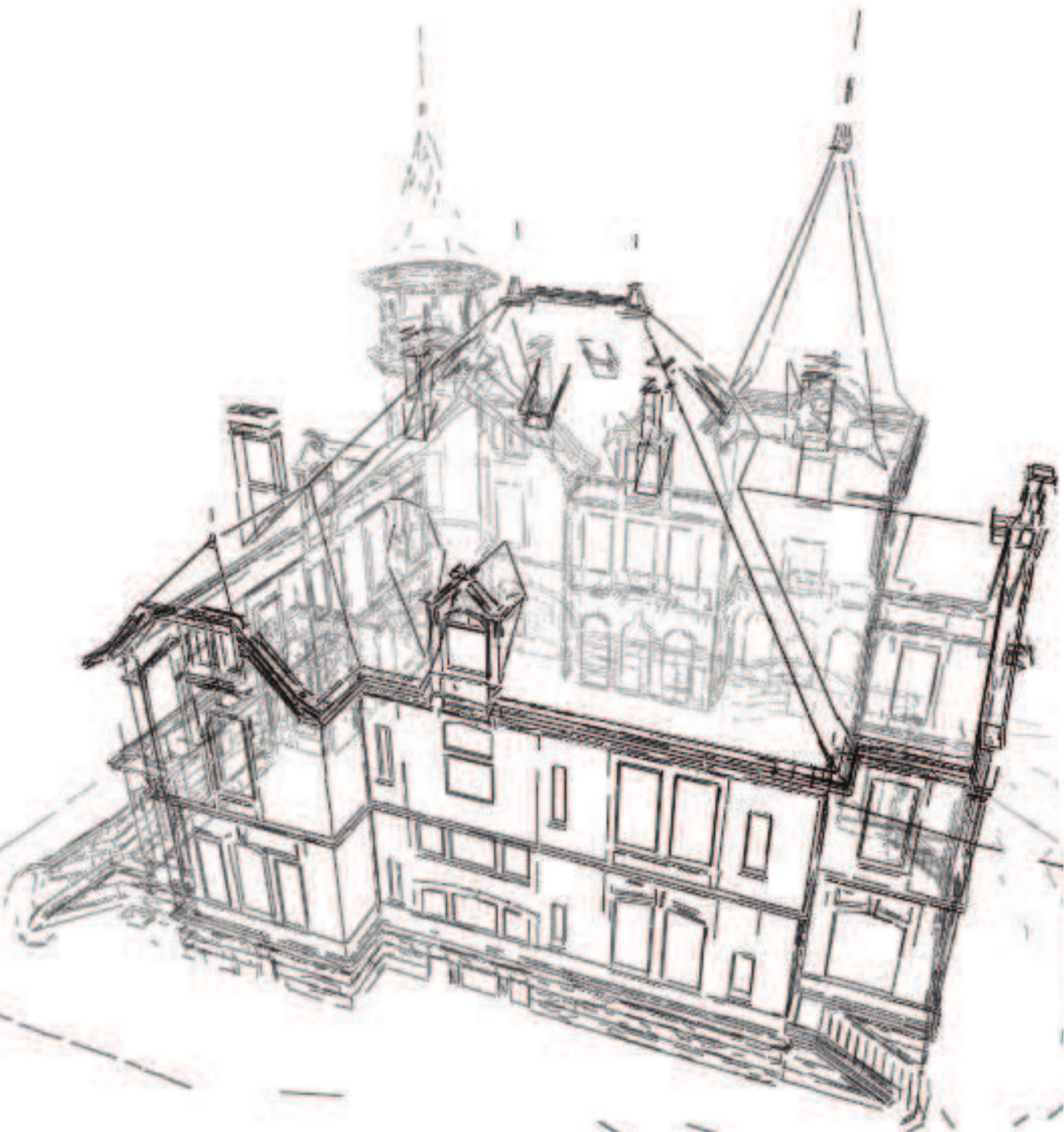} & 
		\includegraphics[height=0.3\linewidth]{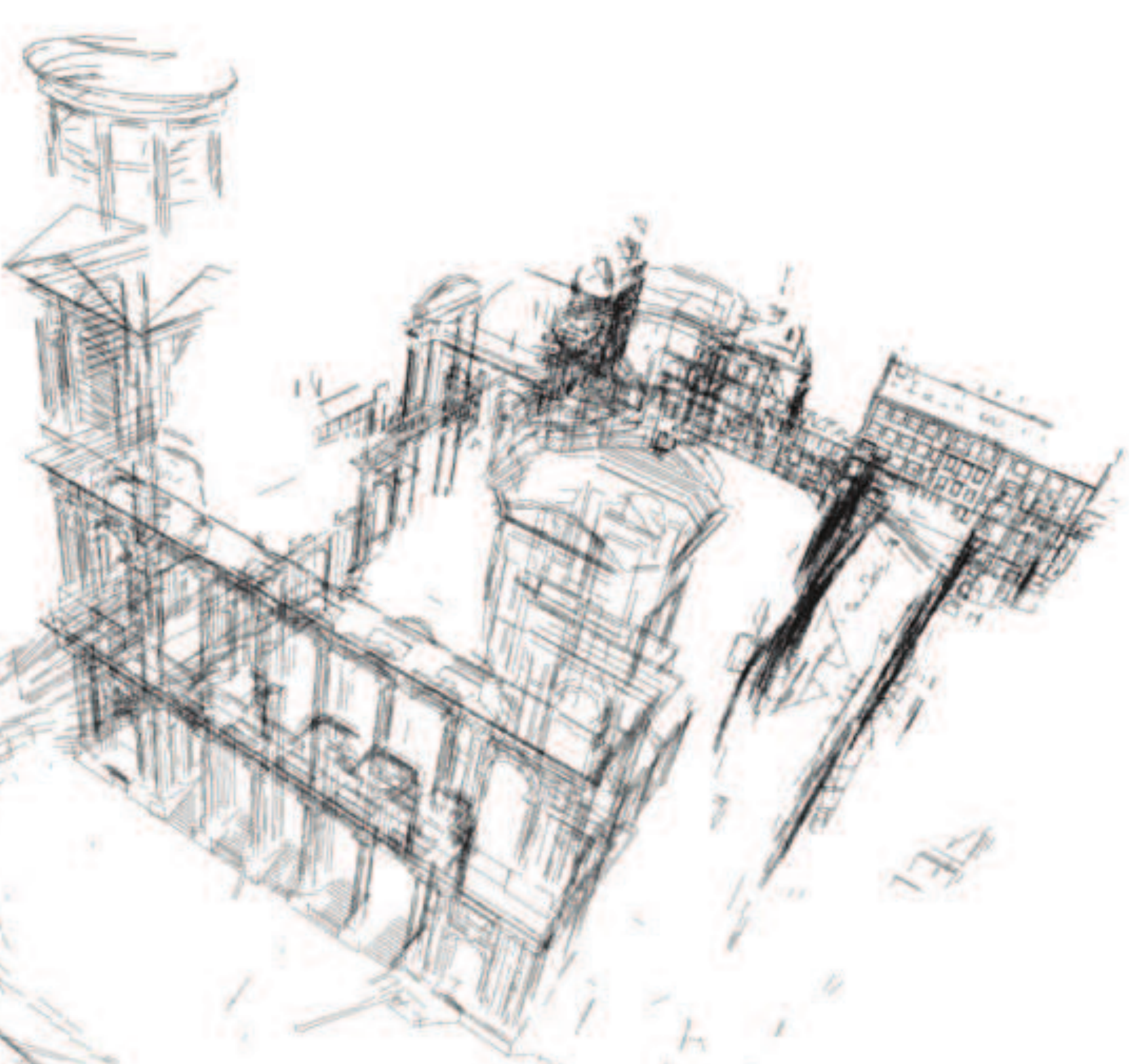} \\
		1104 lines, 37.1s  & 21412 lines, 543.0s  & 8816 lines, 118.6s  \\
		\includegraphics[height=0.3\linewidth]{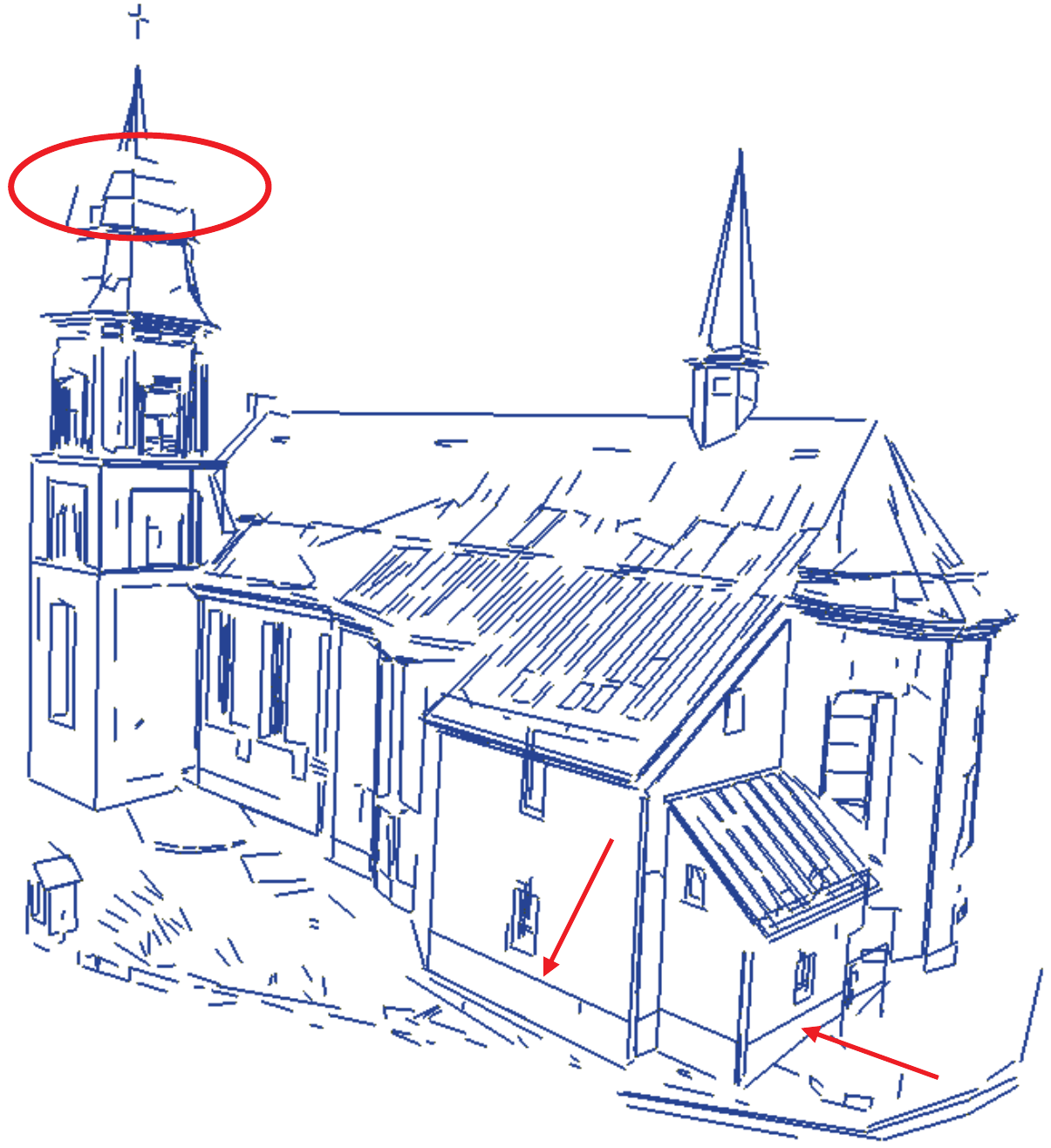} & 
		\includegraphics[height=0.3\linewidth]{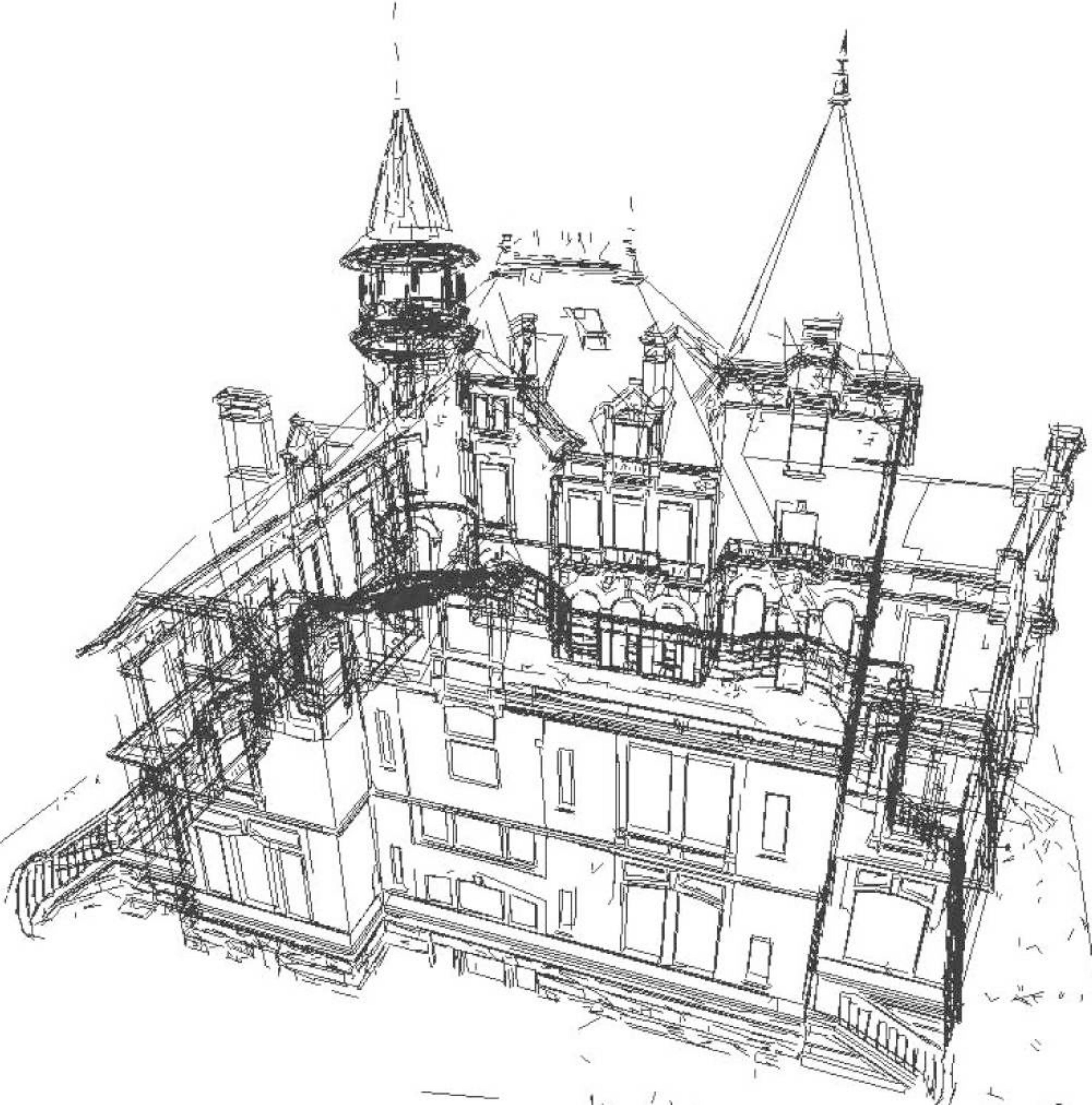} & 
		\includegraphics[height=0.3\linewidth]{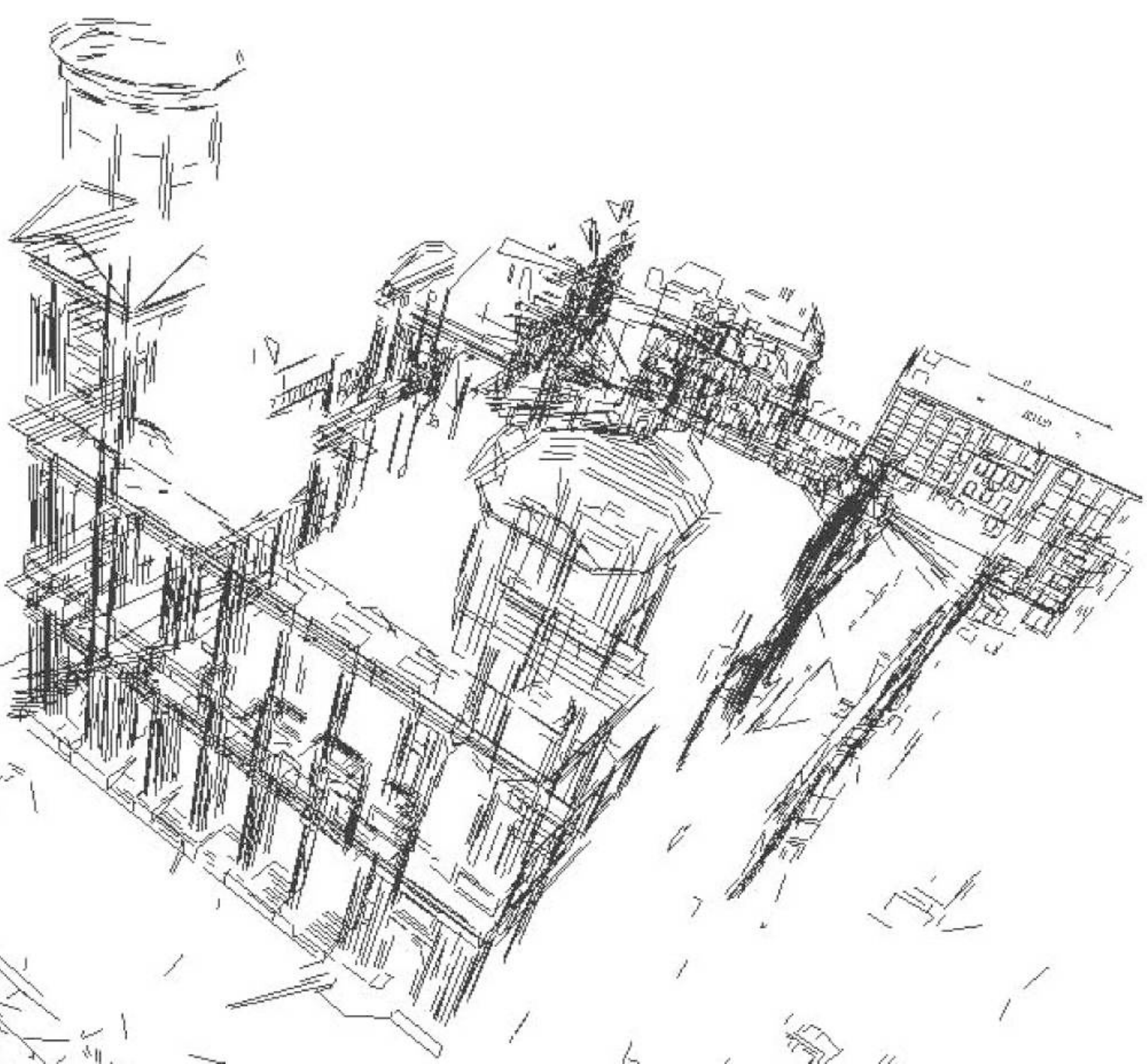} \\
		993 lines, 8.7s    & 12376 lines, 150.1s   & 5450 lines, 30.1s  \\		
	\end{tabular}   	
   	\caption{3D line segment detection results of our method and Lin et al.'s method~\cite{Lin2017Facet} in three public point clouds. First row: results of Lin et al.'s method (obtained from the original article). Second row: results of our proposed method. }
   	\label{Fig:comprison}
\end{figure*} 

\textbf{\textit{Comparison with Other Methods}}: We compared our method with the recent work of Lin et al.~\cite{Lin2017Facet}. Considering that the dataset and source code of Lin's work are not publicly available, we utilized the statistic data and images from their original article. What should be noticed is that some of the tested point clouds were cropped in Lin's experiments, so there may be some small difference between the results of Lin's method and our algorithm. Three public point clouds were tested, including the Bildstein1 (cropped), the Castle (original) and the StSulpice (cropped)~\footnote{Available at \url{http://www.libe57.org/data.html}}. Table~\ref{Tab:comprison} shows the statistic results of the two methods on these point clouds. We can see from Table~\ref{Tab:comprison} that our method is much more efficient than Lin's work in both point cloud segmentation and 3D line detection. Generally, the computing time of our method counts around $20\% - 30\%$ of that of Lin's work. Despite the fact that our method is much more efficient than Lin's work, our detection result is in the same level as that of Lin's work. Figure~\ref{Fig:comprison} show the corresponding visual results of both methods. Along with the number of lines in Table~\ref{Tab:comprison}, we can see that even though our method extracts less lines than Lin's work, the amount of structure in the scene recovered by our method are in the same level as that of Lin's work. The difference between our method and Lin's work lies on that Lin's work can preserve the small line segments on the curve well because it's based on edge points (see the line segments indicated by red ellipses in the first column of Figure~\ref{Fig:comprison}), while our method is plane based so it can recover the structural line segments much better (see the line segments indicated by red arrows in the first column of Figure~\ref{Fig:comprison}).

\textbf{\textit{Failure Cases}}: As we have introduced above, our method is plane based, which is better for structural line segments detection but may also lead to failure in the case of curved facade. The red ellipses in the first column of Figure~\ref{Fig:comprison} indicate the failure cases of our method. Despite these failure cases, our method is very efficient and effective in the man-made scene with lots of structural lines.

\section{Conclusion}
\label{Sec:Conclusion}
In this paper we present and testify a simple but efficient 3D line detection algorithm for large scale unorganized point cloud. The proposed method is based on point cloud segmentation and 2D line detection, also a post-processing procedure is applied to get rid of outliers. Unlike traditional edge point based methods, the proposed method is very easy to be implemented and fast (40s for 10M points). Since our method is based on point cloud segmentation, one failure case of our method is on the curved surface, on which the segmented planes are not real in the scene. Despite this, the proposed method is very efficient for structural scene. In the future, we will make some improvement for the curved surfaces.



%
\bibliographystyle{IEEEtran}
\bibliography{refs}

\end{document}